\def\year{2019}\relax
\documentclass[letterpaper]{article} 
\usepackage{aaai19}  
\usepackage{times}  
\usepackage{helvet}  
\usepackage{courier}  
\usepackage{url}  
\usepackage{graphicx}  
\usepackage{latexsym}
\usepackage{amsmath,amssymb}
\usepackage{bm}
\usepackage{todonotes}
\usepackage{arydshln}
\usepackage{multirow}
\newcommand{\citet}[1]
{\citeauthor{#1}~\shortcite{#1}}
\newcommand{\citep}{\cite}

\begin{document}

%
\title{Zero-Shot Cross-lingual Classification \\
Using Multilingual Neural Machine Translation}

\author{Akiko Eriguchi$^\dagger$ Melvin Johnson$^\ddagger$ Orhan Firat$^\ddagger$ Hideto Kazawa$^\ddagger$ Wolfgang Macherey$^\ddagger$\\
$^\dagger$The University of Tokyo~~~~~~$^\ddagger$Google AI\\
$^\dagger${\tt eriguchi@logos.t.u-tokyo.ac.jp}\\
$^\ddagger${\tt \{melvinp, orhanf, kazawa, wmach\}@google.com}
}

\maketitle
\begin{abstract}
Transferring representations from large supervised tasks to downstream tasks has shown promising results in AI fields such as Computer Vision and Natural Language Processing (NLP). In parallel, the recent progress in Machine Translation (MT) has enabled one to train multilingual Neural MT (NMT) systems that can translate between multiple languages and are also capable of performing zero-shot translation. However, little attention has been paid to leveraging representations learned by a multilingual NMT system to enable zero-shot multilinguality in other NLP tasks. In this paper, we demonstrate a simple framework, a \textit{multilingual Encoder-Classifier}, for cross-lingual transfer learning by reusing the encoder from a multilingual NMT system and stitching it with a task-specific classifier component. Our proposed model achieves significant improvements in the English setup on three benchmark tasks - Amazon Reviews, SST and SNLI. Further, our system can perform classification in a new language for which no classification data was seen during training, showing that zero-shot classification is possible and remarkably competitive. In order to understand the underlying factors contributing to this finding, we conducted a series of analyses on the effect of the shared vocabulary, the training data type for NMT, classifier complexity, encoder representation power, and model generalization on zero-shot performance. Our results provide strong evidence that the representations learned from multilingual NMT systems are widely applicable across languages and tasks.
\end{abstract}

\section{Introduction}
Transfer learning has been shown to work well in Computer Vision where pre-trained components from a model trained on ImageNet~\cite{NIPS2012_4824} are used to initialize models for other tasks~\cite{NIPS2014_5347}. In most cases, the other tasks are related to and share architectural components with the ImageNet task, enabling the use of such pre-trained models for feature extraction. With this transfer capability, improvements have been obtained on other image classification datasets, and on other tasks such as object detection, action recognition, image segmentation, etc~\cite{DBLP:journals/corr/HuhAE16}. Analogously, we propose a method to transfer a pre-trained component - the multilingual encoder from an NMT system - to other NLP tasks.

In NLP, initializing word embeddings with pre-trained word representations obtained from Word2Vec~\cite{NIPS2013_5021} or GloVe~\cite{pennington2014glove} has become a common way of transferring information from large unlabeled data to downstream tasks. Recent work has further shown that we can improve over this approach significantly by considering representations in context, i.e. modeled depending on the sentences that contain them, either by taking the outputs of an encoder in MT~\cite{NIPS2017_7209} or by obtaining representations from the internal states of a bi-directional Language Model (LM)~\cite{elmo}. There has also been successful recent work in transferring sentence representations from resource-rich tasks to improve resource-poor tasks~\cite{D17-1070}, however, most of the above transfer learning examples have focused on transferring knowledge across tasks for a single language, in English.

Cross-lingual or multilingual NLP, the task of transferring knowledge from one language to another, serves as a good test bed for evaluating various transfer learning approaches. For cross-lingual NLP, the most widely studied approach is to use multilingual embeddings as features in neural network models. However, research has shown that representations learned in context are more effective~\cite{NIPS2017_7209,elmo}; therefore, we aim at doing better than just using multilingual embeddings in the cross-lingual tasks. Recent progress in multilingual NMT provides a compelling opportunity for obtaining contextualized multilingual representations, as multilingual NMT systems are capable of generalizing to an unseen language direction, i.e. zero-shot translation. There is also evidence that the encoder of a multilingual NMT system learns language agnostic, universal interlingua representations, which can be further exploited~\cite{TACL1081}.

In this paper, we focus on using the representations obtained from a multilingual NMT system to enable cross-lingual transfer learning on downstream NLP tasks. Our contributions are three-fold:
\begin{enumerate}
    \item We show that by simply reusing the encoder of a multilingual NMT system trained to translate from English to French (En$\rightarrow$Fr) and from French to English (Fr$\rightarrow$En), we can significantly improve the performance over the baseline in three downstream tasks - Amazon Reviews, Stanford Sentiment Treebank (SST), and Stanford Natural Language Inference (SNLI).
    \item We demonstrate that our approach is able to perform zero-shot classification, i.e. performing classification in a new language, e.g. French, without ever seeing any French classification data during training. Our approach obtains surprisingly high zero-shot classification accuracy in French on all three tasks.
    \item We carefully analyze how and why cross-lingual knowledge transfer works, and study the effect of various factors on zero-shot classification performance.
\end{enumerate}

\section{Related Work}
\paragraph{Word and Sentence Representations.}
Pre-trained word representations, which leverage large scale unlabeled data~\cite{NIPS2013_5021,pennington2014glove}, have been shown to be a key ingredient in many standard NLP tasks. The tasks include sentiment analysis~\cite{socher-EtAl:2013:EMNLP}, entailment~\cite{D15-1075}, summarization~\cite{summarization}, question answering~\cite{QA_paper}, and semantic role labeling~\cite{P17-1044}. However, these representations are usually learned from unsupervised data sources which are often unrelated to the downstream task.

\paragraph{Contextualized Representations.}
Several studies have overcome the fact that these representations are context-independent by proposing contextualized word embeddings. Representations obtained from an LM have been shown to obtain effective contextualized word representations~\cite{P17-1161,elmo}. There has also been work in enriching these word representations using sub-word information~\cite{wieting-EtAl:2016:EMNLP2016,Q17-1010}. MT naturally lends itself as a suitable task for obtaining contextualized embeddings since the encoder has to encode units in context so as to decode them into another language.
In \citet{Hill:2017:RGW:3127662.3127714}, the authors show the effectiveness of representations obtained from an NMT model in semantic similarity tasks. They further report that the representations obtained from the NMT model are better than those obtained from LMs. \citet{NIPS2017_7209} showed that using the representations obtained from the encoder of an NMT system as context vectors in downstream NLP tasks significantly improves performance over using only unsupervised word or character $n$-gram vectors.

Finally, there has been a large body of work on obtaining transferable sentence representations. In \citet{D17-1070}, the authors obtain representations from the supervised SNLI task and show that these are effective for transferring to other tasks. Their method outperforms other similar approaches to obtain representations like FastSent~\cite{hill-cho-korhonen:2016:N16-1} and SkipThought~\cite{NIPS2015_5950}. Finally, \citet{arora2016simple} show that a simple average of word embeddings approach is competitive with more complex methods like SkipThought representations.

\paragraph{Cross-lingual or Multilingual Representations.}
Previous approaches to cross-lingual or multilingual representations have fallen into three categories. Obtaining representations from \textit{word level alignments} - bilingual dictionaries or automatically generated word alignments - is the most popular approach~\cite{NIPS2013_5021,faruqui-dyer:2014:EACL,D13-1141}. The second category of methods try to leverage \textit{document level alignment} like parallel Wikipedia articles to generate cross-lingual representations~\cite{sogaard-EtAl:2015:ACL-IJCNLP,Vulic:2016:BDW:3013558.3013583}.

The final category of methods often use \textit{sentence level alignments} in the form of parallel translation data to obtain cross-lingual representations. \citet{hermann-blunsom:2014:P14-1} propose a deep neural model named BiCVM which compares two sentence representations at the final layer and forces them into the same intermediate sentence representation. BilBOWA~\cite{Gouws:2015:BFB:3045118.3045199} is a simpler model which extends skip-gram with negative sampling~\cite{NIPS2013_5021} to optimize each word's similarity with its context in both the current language and the other parallel language. \citet{luong-pham-manning:2015:VSM-NLP} also proposed obtaining cross-lingual representations using a similar approach. \citet{DBLP:journals/corr/AmmarMTLDS16} propose two algorithms, multiCluster and multiCCA, for learning multilingual representations from a set of bilingual lexical data.

Our paper aims to combine the best of both worlds by learning contextualized representations which are multilingual in nature. We demonstrate that using the encoder from a multilingual NMT system as a pre-trained component in other downstream NLP tasks improves performance in English and also enables cross-lingual transfer learning in French.

\section{Proposed Method}
\label{sec:proposed_method}
We propose an \textit{Encoder-Classifier} model, where the \textit{Encoder}, leveraging the representations learned by a multilingual NMT model, converts an input sequence ${\bm x}$ into a set of vectors \textbf{C}, and the \textit{Classifier} predicts a class label $y$ given the encoding of the input sequence, \textbf{C}.

\subsection{Multilingual Representations Using NMT} \label{sec: multirep}
Although there has been a large body of work in building multilingual NMT models which can translate between multiple languages at the same time~\cite{LuongLSVK15,dong2015multi,firat2016multi,TACL1081}, zero-shot capabilities
of such multilingual representations have only been tested for MT~\cite{TACL1081}. 
We propose a simple yet effective solution - reuse the encoder of a multilingual NMT model to initialize the encoder for other NLP tasks. 
To be able to achieve promising zero-shot classification performance, we consider two factors: 
(1) The ability to encode multiple source languages with the same encoder and (2) The ability to learn language agnostic representations of the source sequence. Based on the literature, both requirements can be satisfied by training a multilingual NMT model having a shared encoder~\cite{DBLP:journals/corr/LeeCH16,TACL1081}, and a separate decoder and attention mechanism for each target language~\cite{dong2015multi}. After training such a multilingual NMT model, the decoder and the corresponding attention mechanisms (which are target-language specific) are discarded, while the multilingual encoder is used to initialize the encoder of our proposed \textit{Encoder-Classifier} model.

\subsection{Multilingual Encoder-Classifier}

\paragraph{Encoder}
In order to leverage pre-trained multilingual representations introduced in Section~\ref{sec: multirep}, our encoder strictly follows the structure of a regular Recurrent Neural Network (RNN) based NMT encoder~\cite{BahdanauCB15}
with a stacked layout~\cite{DBLP:journals/corr/WuSCLNMKCGMKSJL16}. Given an input sequence 
${\bm x} = (x_{1}, x_{2}, \ldots, x_{T_x})$ of length $T_x$, our encoder contextualizes or encodes the input sequence
into a set of vectors \textbf{C}, by first applying a bi-directional RNN~\cite{Schuster:1997:BRN:2198065.2205129}, followed by a stack of uni-directional RNNs. The hidden states of the final layer RNN, $h_i^l$, form the set \textbf{C}$~=\{ h_i^l \}_{i=1}^{T_x}$ of context vectors which will be used by the classifier, where $l$ denotes the number of RNN layers in the stacked encoder.

\paragraph{Classifier} 
The task of the classifier is to predict a class label $y$ given the context set \textbf{C}. To ease this classification task given a variable length input set \textbf{C}, a common approach in the literature is to extract a single sentence vector $\bm{q}$ by making use of pooling over time~\cite{journals/jmlr/CollobertWBKKK11}.
Further, to increase the modeling capacity, the pooling operation can be parameterized using pre- and post-pooling networks. Formally, given the context set \textbf{C}, we extract a sentence vector $\bm{q}$ in three steps, 
using three networks, (1) pre-pooling feed-forward network $f_{pre}$, (2) pooling network $f_{pool}$ and (3) post-pooling feed-forward network $f_{post}$,
\begin{align*}
\bm{q} = f_{post}( f_{pool} ( f_{pre} (\textbf{C})  )  ).
\end{align*}
Finally, given the sentence vector $\bm{q}$, a class label $y$ is predicted by employing a softmax function.

\section{Experimental Design}

\subsection{Corpora}
We evaluate the proposed method on three common NLP tasks: Amazon Reviews, SST and SNLI. We utilize parallel data to train our multilingual NMT system, as detailed below. 

\paragraph{Machine Translation}  \label{subsec:mt}
For the MT task, we use the WMT 2014 En$\leftrightarrow$Fr parallel corpus. The dataset contains 36 million En$\rightarrow$Fr sentence pairs. We swapped the source and target sentences to obtain parallel data for the Fr$\rightarrow$En translation task. We use these two datasets (72 million sentence pairs) to train a single multilingual NMT model to learn both these translation directions simultaneously. We generated a shared sub-word vocabulary~\cite{SennrichHB15,conf/icassp/SchusterN12} of 32K units from all source and target training data. We use this sub-word vocabulary for all of our experiments below.

\paragraph{Amazon Reviews}
The Amazon reviews dataset~\cite{Prettenhofer:2010:CTC:1858681.1858795} is a multilingual sentiment classification dataset, providing data for four languages - English (En), French (Fr), German (De), and Japanese. We use the English and French datasets in our experiments. The dataset contains 6,000 documents in the train and test portions for each language. Each review consists of a category label, a title, a review, and a star rating (5-point scale). We only use the review text in our experiments. Following~\citet{Prettenhofer:2010:CTC:1858681.1858795}, we mapped the reviews with lower scores (1 and 2) to negative examples and the reviews with higher scores (4 and 5) to positive examples, thereby turning it into a binary classification problem. Reviews with score 3 are dropped. We split the training dataset into 10\% for development and the rest for training, and we truncate each example and keep the first 200 words in the review. Note that, since the data for each language was obtained by crawling different product pages, the data is not aligned across languages.

\paragraph{SST} \label{sec: test data sst}
The sentiment classification task proposed in~\citet{socher-EtAl:2013:EMNLP} is also a binary classification problem where each sentence and phrase is associated with either a positive or a negative sentiment. We ignore phrase-level annotations and sentence-level neutral examples in our experiments. The dataset contains 6920, 872, and 1821 examples for training, development and testing, respectively. Since SST does not provide a multilingual test set, we used the public translation engine Google Translate\footnote{\url{https://translate.google.com} as of October 2017.} to translate the SST test set to French. Previous work by \citet{lrec2018} has shown that replacing the human translated test set with a synthetic set (obtained by using Google Translate) produces only a small difference of around 1\% absolute accuracy on their human-translated French SNLI test set. Therefore, the performance measured on our `pseudo' French SST test set is expected to be a good indicator of zero-shot performance.

\paragraph{Multilingual SNLI}
Natural language inference is a task that aims to determine whether a natural language hypothesis $\bm{h}$ can justifiably be inferred from a natural language premise $\bm{p}$. SNLI~\cite{D15-1075} is one of the largest datasets for a natural language inference task in English and contains multiple sentence pairs with a sentence-level entailment label. Each pair of sentences can have one of three labels - {\it entailment}, {\it contradiction}, and {\it neutral}, which are annotated by multiple humans. The dataset contains 550K training, 10K validation, and 10K testing examples. To enable research on multilingual SNLI, \citet{lrec2018} chose a subset of the SNLI test set (1332 sentences) and professionally translated it into four major languages - Arabic, French, Russian, and Spanish. We use the French test set for evaluation in Section \ref{sec: zero-shot_classification} and \ref{sec: analyses}.

\begin{table*}[!ht]
\begin{center}
\begin{tabular}{| l | c c | c | c |}
\hline \bf Model & Amazon (En) &  Amazon (Fr) & SST (En) & SNLI (En) \\ \hline
Proposed model: \textit{Encoder-Classifier}      & 76.60 & 82.50 & 79.63 & 76.70\\
+ Pre-trained Encoder	& 80.70 & 83.18 & 84.18 & 84.42\\
+ Freeze Encoder 		& 84.13 & 85.65 & 84.51 & 84.41\\ \hdashline
State-of-the-art Models 
                        & 83.50  & 87.50 & 90.30 & 88.10 \\
\hline
\end{tabular}
\end{center}
\caption{\label{tab: results_all} Transfer learning results of the classification accuracy on all the datasets. Amazon (En) and Amazon (Fr) are the English and French versions of the task, training the models on the data for each language. The state-of-the-art results are cited from~\citet{ijcai2018-802} for both Amazon Reviews tasks and~\citet{NIPS2017_7209} for SST and SNLI.
}
\end{table*}

\subsection{Model and Training Details}
\label{sec:training_details}
Here, we first describe the model and training details of the base multilingual NMT model
whose encoder is reused in all other tasks. Then we provide details about  
the task-specific classifiers. For each task, we provide the specifics of 
$f_{pre}$, $f_{pool}$ and $f_{post}$ nets that build the task-specific classifier.

All the models in our experiments are trained using Adam optimizer~\cite{journals/corr/KingmaB14} with label smoothing 
~\cite{DBLP:conf/cvpr/SzegedyVISW16} and unless otherwise stated below, 
layer normalization~\cite{DBLP:journals/corr/BaKH16} is applied to all LSTM gates and feed-forward layer inputs. We apply L2 regularization to the model weights and dropout to layer activations and sub-word embeddings.
Hyper-parameters, such as mixing ratio $\lambda$ of L2 regularization, dropout rates, label smoothing uncertainty, batch sizes, 
learning rate of optimizers and initialization ranges of weights are tuned on the development sets provided for each task
separately.

\paragraph{NMT Models} 
\label{subsec:NMT models}
Our multilingual NMT model consists of a shared multilingual encoder and two decoders, one for English and the other for French. The multilingual encoder uses one bi-directional LSTM, followed by three stacked layers of uni-directional LSTMs in the encoder. Each decoder consists of four stacked LSTM layers, with the first LSTM layers intertwined with additive attention networks~\cite{BahdanauCB15} to learn a source-target alignment function. All the uni-directional LSTMs are equipped with residual connections~\cite{DBLP:conf/cvpr/HeZRS16} to ease the optimization, both in the encoder and the decoders.
LSTM hidden units and the shared source-target embedding dimensions are set to 512.

Similar to~\citet{dong2015multi}, multilingual NMT model is trained in a multi-task learning setup, where each decoder is augmented with a task-specific loss, minimizing the negative conditional log-likelihood of the target sequence given the source sequence. During training, mini-batches of En$\rightarrow$Fr and Fr$\rightarrow$En examples are interleaved. We picked the best model based on the best average development set BLEU score on both of the language pairs.

\paragraph{Amazon Reviews and SST} \label{subsec:amazon}
The \textit{Encoder-Classifier} model here uses the encoder defined previously. With regards to the classifier, the pre- and post-pooling networks ($f_{pre}$,  $f_{post}$) are both one-layer feed forward networks to cast the dimension size from 512 to 128 and from 128 to 32, respectively. We used max-pooling operator for the $f_{pool}$ network to pool the activation over time.

\paragraph{Multilingual SNLI}

We extended the proposed \textit{Encoder-Classifier} model to a multi-source  model~\cite{DBLP:journals/corr/ZophK16} since SNLI is an inference task of relations between two input sentences, \lq\lq premise" and \lq\lq hypothesis". For the two sources, we use two separate encoders, which are initialized with the same pre-trained multilingual NMT encoder, to obtain 
their representations. Following our notation, the encoder outputs are processed using $f_{pre}$, $f_{pool}$ and $f_{post}$ nets, again with two separate network blocks. Specifically, $f_{pre}$ consists of a co-attention layer~\cite{DBLP:journals/corr/LuYBP16} followed by a two-layer feed-forward neural network with residual connections. 
We use max pooling over time for $f_{pool}$ and again a two-layer feed-forward neural network with residual connections as $f_{post}$. After processing two sentence encodings using two network blocks, 
we obtain two vectors representing premise $\bm{h}_{premise}$ and hypothesis $\bm{h}_{hypothesis}$.
Following~\citet{tai-socher-manning:2015:ACL-IJCNLP}, we compute two types of relational vectors with $\bm{h}_{-} =  |\bm{h}_{premise} - \bm{h}_{hypothesis}|,$ and $\bm{h}_{\times} = \bm{h}_{premise} \odot \bm{h}_{hypothesis}$, where $\odot$ denotes the element-wise multiplication between two vectors. The final relation vector is obtained by concatenating $\bm{h}_{-}$ and $\bm{h}_{\times}$. For both \lq\lq premise" and \lq\lq hypothesis" feed-forward networks we used 512 hidden dimensions. 

For Amazon Reviews, SST and SNLI tasks, we picked the best model based on the highest development set accuracy.

\begin{table*}[t!]
\begin{center}
\begin{tabular}{| l | c c | c c | c c |}
\hline \bf \multirow{2}{*}{Model} & \multicolumn{2}{c|}{Amazon (Fr)} & \multicolumn{2}{c|}{SST (Fr)} & \multicolumn{2}{c|}{SNLI (Fr)}\\ \cline{2-7} 
& Bridged & Zero-Shot & Bridged$^*$ & Zero-Shot & Bridged & Zero-Shot\\ \hline
Proposed model: \textit{Encoder-Classifier} & 73.30 & 51.53 & 79.63 & 59.47 & 74.41 & 37.62\\
+ Pre-trained Encoder	& 79.23 & 75.78 & 84.18 & 81.05 & 80.65 & 72.35\\
+ Freeze Encoder 		& 83.10 & 81.32 & 84.51 & 83.14 & 81.26 & 73.88\\
\hline
\end{tabular}
\end{center}
\caption{\label{table: zero_shot} Zero-Shot performance on all French test sets. $^*$Note that we use the English accuracy in the bridged column for SST.}
\end{table*}

\section{Transfer Learning Results}
In this section, we report our results for the three tasks - Amazon Reviews (English and French), SST, and SNLI. For each task, we first build a baseline system using the proposed \textit{Encoder-Classifier} architecture described in Section \ref{sec:proposed_method} where the encoder is initialized randomly. Next, we experiment with using the pre-trained multilingual NMT encoder to initialize the system as described in Section~\ref{sec: multirep}. Finally, we perform an experiment where we freeze the encoder after initialization and only update the classifier component of the system.

Table~\ref{tab: results_all} summarizes the accuracy of our proposed system for these three different approaches and the state-of-the-art results on all the tasks. The first row in the table shows the baseline accuracy of our system for all four datasets. The second row shows the result from initializing with a pre-trained multilingual NMT encoder. It can be seen that this provides a significant improvement in accuracy, an average of 4.63\%, across all the tasks. This illustrates that the multilingual NMT encoder has successfully learned transferable contextualized representations that are leveraged by the classifier component of our proposed system. These results are in line with the results in~\citet{NIPS2017_7209} where the authors used the representations from the top NMT encoder layer as an additional input to the task-specific system. However, in our setup we reused all of the layers of the encoder as a single pre-trained component in the task-specific system. The third row shows the results from freezing the pre-trained encoder after initialization and only training the classifier component. For the Amazon English and French tasks, freezing the encoder after initialization significantly improves the performance further. We hypothesize that since the Amazon dataset is a document level classification task, the long input sequences are very different from the short sequences consumed by the NMT system and hence freezing the encoder seems to have a positive effect. This hypothesis is also supported by the SNLI and SST results, which contain sentence-level input sequences, where we did not find any significant difference between freezing and not freezing the encoder.

\section{Zero-Shot Classification Results}
\label{sec: zero-shot_classification}
In this section, we explore the zero-shot classification task in French for our systems. We assume that we do not have any French training data for all the three tasks and test how well our proposed method can generalize to the unseen French language without any further training. Specifically, we reuse the three proposed systems from Table~\ref{tab: results_all} after being trained only on the English classification task and test the systems on data from an unseen language (e.g. French). A reasonable upper bound to which zero-shot performance should be compared to is \textit{bridging} - translating a French test text to English and then applying the English classifier on the translated text. If we assume the translation to be perfect, we should expect this approach to perform as well as the English classifier.

The Amazon Reviews and SNLI tasks have a French test set available, and we evaluate the performance of the bridged and zero-shot systems on each French set. However, the SST dataset does not have a French test set, hence the `pseudo French' test set described in Section~\ref{sec: test data sst} is used to evaluate the zero-shot performance. We use the English accuracy scores from the SST column in Table~\ref{tab: results_all} as a high-quality proxy for the SST bridged system. We do this since translating the `pseudo French' back to English will result in two distinct translation steps and hence more errors.

Table~\ref{table: zero_shot} summarizes all of our zero-shot results for French classification on the three tasks. It can be seen that just by using the pre-trained NMT encoder, the zero-shot performance increases drastically from almost random to within 10\% of the bridged system. 
Freezing the encoder further pushes this performance closer to the bridged system. On the Amazon Review task, our zero-shot system is within 2\% of the best bridged system. On the SST task, our zero-shot system obtains an accuracy of 83.14\% which is within 1.5\% of the bridged equivalent (in this case the English system).

Finally, on SNLI, we compare our best zero-shot system with bilingual and multilingual embedding based methods evaluated on the same French test set in~\citet{lrec2018}. As illustrated in Table~\ref{table: compare_zeroshot}, our best zero-shot system obtains the highest accuracy of 73.88\%. INVERT~\cite{sogaard-EtAl:2015:ACL-IJCNLP} uses inverted indexing over a parallel corpus to obtain crosslingual word representations. BiCVM~\cite{hermann-blunsom:2014:P14-1} learns bilingual compositional representations from sentence-aligned parallel corpora. In RANDOM~\cite{Vulic:2016:BDW:3013558.3013583}, bilingual embeddings are trained on top of parallel sentences with randomly shuffled tokens using skip-gram with negative sampling, and RATIO is similar to RANDOM with the one difference being that the tokens in the parallel sentences are not randomly shuffled. Our system significantly outperforms all methods listed in the second column by 10.66\% to 15.24\% and demonstrates the effectiveness of our proposed approach.

\begin{table}[t!]
\begin{center}
\begin{tabular}{| l | c | c |}
\hline  	
\bf Model & SNLI (Fr) \\ \hline
Our best zero-shot \textit{Encoder-Classifier} & \textbf{73.88}\\\hdashline
INVERT~\cite{sogaard-EtAl:2015:ACL-IJCNLP}	 & 62.60\\
BiCVM~\cite{hermann-blunsom:2014:P14-1}		 & 59.03\\
RANDOM~\cite{Vulic:2016:BDW:3013558.3013583} & 63.21\\
RATIO~\cite{Vulic:2016:BDW:3013558.3013583}	 & 58.64\\\hline
\end{tabular}
\end{center}
\caption{\label{table: compare_zeroshot} Comparison of our best zero-shot result on the French SNLI test set to other baselines. See text for details.}
\end{table}

\section{Analyses}
\label{sec: analyses}
In this section, we try to analyze why our simple \textit{Encoder-Classifier} system is effective at zero-shot classification. We perform a series of experiments to better understand this phenomenon. In particular, we study (1) the effect of shared sub-word vocabulary, (2) the amount of multilingual training data to measure the influence of multilinguality, (3) encoder/classifier capacity to measure the influence of representation power, and (4) model behavior on different training phases to assess the relation between generalization performance on English and zero-shot performance on French.

\paragraph{Effect of Shared Sub-Word Vocabulary.}
As mentioned in Section~\ref{sec:training_details}, we use a shared sub-word vocabulary which can encode both English and French text in all of our models. In this subsection, we analyze how much using a shared sub-word vocabulary can help the model generalize to a new language. To verify the effectiveness of just the sub-word vocabulary on generalization, we picked the German test set from the Amazon Review task. Since German shares many sub-words with English and French, the Out-Of-Vocabulary (OOV) rate for the German test set using our vocabulary is just 0.078\%. 
We design this experiment as a control to understand the effect of having a shared sub-word vocabulary 
which can encode the language but for which no translation data was seen while training the multilingual NMT encoder.

\begin{table}[!t]
\begin{center}
\begin{tabular}{| l | c |}
\hline  	
\bf Model			& Amazon (De) \\ \hline
Zero-shot \textit{Encoder-Classifier}	&52.33  	\\
+ Pre-trained Encoder			& 52.98	\\
+ Freeze Encoder				&  57.72	\\\hline
\end{tabular}
\end{center}
\caption{\label{table:german_zeroshot} Results of the control experiment on zero-shot performance on the Amazon German test set. }
\end{table}

From Table~\ref{table:german_zeroshot}, we can see that despite the very low OOV rate, the ability of our system to perform zero-shot classification on German is close to random, i.e. around 50\% accuracy. The third row in the table shows the small deviation of 7\% over random, which is likely obtained from common sub-words having similar meaning across languages. This control experiment suggests that although having a shared sub-word vocabulary is necessary, we still need to train the NMT system on parallel data from the language of interest so that the system can perform zero-shot classification.

\paragraph{Effect of Translation Data.}
We explore two dimensions that could affect zero-shot performance related to our training data in the multilingual NMT model. First, we investigate the effect of using symmetric training data to train both directions in the multilingual NMT system. We conduct an experiment where we take half of the sentences from the En$\rightarrow$Fr training set and use the swapped version of the other half of the sentences for training the model. Second, we want to see the effect of training data size, so we run an experiment where we use only half of the training set in a symmetric fashion. From Table~\ref{table:zero_shot_data}, we can see that halving the training data size significantly lowers the zero-shot accuracy on the French SNLI test set by 7.16\%. However, both the symmetric and asymmetric versions of the data perform comparably on both tasks. This shows that the multilingual NMT system is able to learn an effective interlingua even without the need of symmetric data across the language pairs involved.

\begin{table}[t!]
\begin{center}
\begin{tabular}{| l | c c|}
\hline \bf Parallel data type for NMT & SNLI (En) & SNLI (Fr)\\ \hline
Symmetric data (full)   & 84.13 & 73.88 \\ \hdashline
Symmetric data (half)   & 80.79 & 66.72 \\
Asymmetric data (half)  & 81.15 & 67.63 \\
\hline
\end{tabular}
\end{center}
\caption{\label{table:zero_shot_data} Effect of machine translation data over our proposed \textit{Encoder-Classifier} on the SNLI tasks. The results of SNLI (Fr) shows the zero-shot performance of our system. }
\end{table}

\begin{table*}[!t]
\begin{center}
\begin{tabular}{| l | c c | c c|}
\hline \bf \multirow{2}{*}{Encoder components} & \multicolumn{2}{c|}{Simpler classifier} & \multicolumn{2}{c|}{Complex classifier}\\ \cline{2-5}
& SNLI (En) & SNLI (Fr) & SNLI (En) & SNLI (Fr) \\ \hline
Embeddings only & 65.18 & 49.66 & 82.43 & 56.66\\
+ bi-directional layer 1 & 67.99 & 58.19 & 83.40 & 64.74\\
+ layer 2 & 67.00 & 61.01 & 83.63 & 72.81\\
+ layer 3 & 67.26 & 60.55 & 84.17 & 74.33\\
+ layer 4 & 67.26 & 61.61 & 84.41 & 74.11\\ \hline
\end{tabular}
\end{center}
\caption{\label{table:zero_shot_capacity} Zero-shot analyses of classifier network model capacity. The SNLI (Fr) results report the zero-shot performance.}
\end{table*}

\paragraph{Effect of Encoder/Classifier Capacity}
We study the effect of the capacity of the two parts of our model on the final accuracies. Specifically, we experimented with two variants of the classifier - a simple linear classifier where we set $f_{pre}$ and $f_{post}$ networks to identity\footnote{We empirically found that for simple classifiers using mean pooling for $f_{pool}$ performs considerably better over max-pooling (67.26 vs 61.19 test accuracies respectively) on the SNLI task.} and a complex classifier (details provided in Section~\ref{sec:training_details}). Next, we experimented with only reusing different parts of the multilingual encoder in a bottom-up fashion. Table~\ref{table:zero_shot_capacity} summarizes all of our experiments with respect to model capacity. It can be seen that, as expected going from a simple linear classifier to a complex classifier significantly improves both English and zero-shot French performance on the SNLI tasks. However, even a simple linear classifier can achieve significant zero-shot performance when provided with rich enough encodings (49.66 to 61.61 accuracy). However, changing the encoder capacity tells an interesting story. As we selectively reuse parts of the encoder from the embedding layer to the top, we notice that the English performance only increases by about 2\% whereas the zero-shot performance increases by about 18\% at most in the complex classifier. This means that the additional layers in the encoder are essential for the proposed system to model a language agnostic representation (interlingua) which enables it to perform better zero-shot classification. Moreover, it should be noted that best zero-shot performance is obtained by using the complex classifier and up to layer 3 of the encoder. Although this gap is not big enough to be significant, we hypothesize that top layer of the encoder could be very specific to the MT task and hence might not be best suited for zero-shot classification.

\begin{figure}[!t]
\centering
\includegraphics[width=\linewidth]{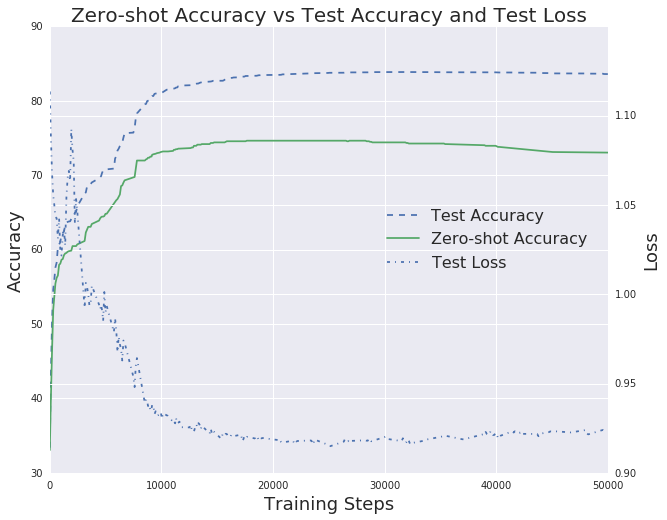}
\caption{\label{fig: zero-shot-graph}Correlation between test-loss, test-accuracy (the English SNLI) and zero-shot accuracy (the French test set).}
\label{fig:fig1}
\end{figure}

\vspace{-5px}
\paragraph{Effect of Early vs Late Phases of the Training}
Figure~\ref{fig: zero-shot-graph} shows that as the number of training steps increases, the test accuracy goes up whereas the test loss on the SNLI task increases slightly hinting at over-fitting on the English task. As expected, choosing checkpoints which are before the onset of the over-fitting seems to benefit zero-shot performance on the French SNLI test set. This suggests that over-training on the English task might hurt the ability of the model to generalize to a new language and also motivated us to conduct the next set of analysis.\footnote{We observe that test loss better correlates with zero-shot accuracy than test accuracy.}

\begin{table}[!t]
\centering
\label{my-label}
\begin{tabular}{|c|cc|}
\hline
Smoothing Range (steps) & SNLI (En)  & SNLI (Fr) \\ \hline
1                       & 84.41 & 74.11     \\
400                     & 84.62 & 75.02     \\
1K                      & 84.67 & 75.48     \\
20K                     & 84.65 & 75.93     \\
35K                     & 84.46 & 75.63     \\ \hline
\end{tabular}
\caption{\label{table: zero_shot_avg_ckpt} Effect of parameter smoothing on the English SNLI test set and zero-shot performance on the French test set.}
\end{table}

\vspace{-5px}
\paragraph{Effect of Parameter Smoothing}
Parameter smoothing (checkpoint averaging~\cite{DBLP:journals/corr/Junczys-Dowmunt16a}) is a technique which aims to smooth point estimates of the learned parameters by averaging $n$ steps from the training run and using it for inference. This is aimed at improving generalization and being less susceptible to the effects of over-fitting at inference. We hypothesize that a system with enhanced generalization might be better suited for zero-shot classification since it is a measure of the ability of the model to generalize to a new task. Table~\ref{table: zero_shot_avg_ckpt} validates our hypothesis by showing that although the average of 20k steps only improves the English SNLI score by 0.24\%, it improves the corresponding French zero-shot score by 1.82\%.

\section{Conclusion}
In this paper, we have demonstrated a simple yet effective approach to perform cross-lingual transfer learning using representations from a multilingual NMT model. Our proposed approach of reusing the encoder from a multilingual NMT system as a pre-trained component provides significant improvements on three downstream tasks. Further, our approach enables us to perform surprisingly competitive zero-shot classification on an unseen language and outperforms cross-lingual embedding base methods. Finally, we end with a series of analyses which shed light on the factors that contribute to the zero-shot phenomenon. We hope that these results showcase the efficacy of multilingual NMT to learn transferable contextualized representations for many downstream tasks.


\bibliographystyle{aaai}
\bibliography{aaai19.bib}
\end{document}